\documentclass[
]{ceurart}

\usepackage{listings}
\begin{document}

\copyrightyear{2025}
\copyrightclause{Copyright for this paper by its authors. Use permitted under Creative Commons License Attribution 4.0 International (CC BY 4.0).}
\conference{5th International Workshop on Scientific Knowledge: Representation, Discovery, and Assessment, Nov 2024, Nara, Japan}

\title{Towards AI-Supported Research: a Vision of the TIB AIssistant}

\author[1,2]{S\"oren Auer}[%
orcid=0000-0002-0698-2864,
email=auer@tib.eu,
]

\author[1]{Allard Oelen}[%
orcid=0000-0001-9924-9153,
email=allard.oelen@tib.eu,
]
\address[1]{TIB – Leibniz Information Centre for Science and Technology, Hannover, Germany}

\author[2,1]{Mohamad Yaser Jaradeh}[%
orcid=0000-0001-8777-2780,
email=jaradeh@l3s.de,
]
\address[2]{L3S Research Center, Leibniz University of Hannover, Hannover, Germany}

\author[1]{Mutahira Khalid}[%
orcid=0000-0001-8482-4004,
email=mutahira.khalid@tib.eu,
]

\author[1]{Farhana Keya}[%
orcid=0000-0000-0000-0000,
email=farhana.keya@tib.eu,
]

\author[1]{Sasi Kiran Gaddipati}[%
orcid=0000-0003-3098-4592,
email=sasi.gaddipati@tib.eu,
]

\author[1]{Jennifer D'Souza}[%
orcid=0000-0002-6616-9509,
email=jennifer.dsouza@tib.eu,
]

\author[3]{Lorenz Schlüter}[%
email=lorenz.schlueter@stud.uni-hannover.de,
]
\address[3]{Leibniz University of Hannover, Hannover, Germany}

\author[3]{Amirreza Alasti}[%
orcid=0009-0002-1165-773X,
email=amirreza.alasti@stud.uni-hannover.de,
]

\author[2]{Gollam Rabby}[%
orcid=0000-0002-1212-0101,
email=gollam.rabby@l3s.de,
]

\author[4,1]{Azanzi Jiomekong}[
orcid=0000-0002-8005-2067,
email=jiomekong@tib.eu,
]
\address[4]{University of Yaounde 1, Yaounde, Cameroon}

\author[1]{Oliver Karras}[%
orcid=0000-0001-5336-6899,
email=oliver.karras@tib.eu,
]

\begin{abstract}
The rapid advancements in Generative AI and Large Language Models promise to transform the way research is conducted, potentially offering unprecedented opportunities to augment scholarly workflows. 
However, effectively integrating AI into research remains a challenge due to varying domain requirements, limited AI literacy, the complexity of coordinating tools and agents, and the unclear accuracy of Generative AI in research. 
We present the vision of the TIB AIssistant, a domain-agnostic human-machine collaborative platform designed to support researchers across disciplines in scientific discovery, with AI assistants supporting tasks across the research life cycle. 
The platform offers modular components — including prompt and tool libraries, a shared data store, and a flexible orchestration framework — that collectively facilitate ideation, literature analysis, methodology development, data analysis, and scholarly writing.  
We describe the conceptual framework, system architecture, and implementation of an early prototype that demonstrates the feasibility and potential impact of our approach. 
\end{abstract}


\maketitle

\section{Introduction}
\label{sec:introduction}
The developments of Generative AI, and specifically Large Language Models (LLMs), have had a significant impact in many areas of our society~\cite{haque2025exploring} already. Additionally, in the scientific domain, LLMs are increasingly utilized, for example, to assist researchers with academic writing~\cite{liang2024mapping}. LLMs are used across a wide variety of scholarly domains, such as medicine in life sciences~\cite{thirunavukarasu2023large}, social sciences~\cite{grossmann2023ai}, chemistry~\cite{branAugmentingLargeLanguage2024}, law~\cite{siino2025exploring}, and coding in computer science~\cite{kazemitabaar2024codeaid}. 

While many approaches have been proposed and demonstrated, it remains challenging for researchers to get started with LLMs in their field. The ability of individual researchers to optimally leverage AI for their research strongly depends on their AI literacy, i.e., their ability to evaluate, communicate with, and collaborate using AI~\cite{long2020ai}. AI can be used to support domain-independent tasks, such as finding related work, assisting with paper authoring, and proofreading, as well as for domain-specific tasks, including supporting methodologies, implementations, or evaluations. While the possibilities are virtually unlimited, the benefits that a single researcher gains from AI-assisted research heavily depend on the user and the specifics of her research work, which determine the required prompts and the tasks that can be outsourced to the LLM. Prompt engineering is a crucial aspect for effective LLM usage~\cite{knoth2024ai}, but can be a bottleneck for non-AI experts~\cite{zamfirescu2023johnny}. Even if researchers possess the necessary skills to operate LLMs, research work relies on diverse tools designed to support specific tasks, such as data analysis in computing environments like R Studio or digital libraries that support knowledge discovery. The effective integration of Generative AI with the diverse tools used in research remains a challenge. For example, the integration of an LLM-based assistant with digital libraries can provide additional context via Retrieval-Augmented Generation~\cite{lewis2020retrieval} or by calling external services using Tool Callings~\cite{schick2023toolformer}.

Based on these considerations, we identified the following challenges for AI-assisted research:
\begin{itemize}
    \item \textbf{Challenge 1}: lacking domain-specific AI literacy to leverage AI for research tasks effectively.
    \item \textbf{Challenge 2}: the skill to effectively engineer prompts and context injection.
    \item \textbf{Challenge 3}: leveraging existing tools and services into AI workflows and providing appropriate user interfaces for them.
    \item \textbf{Challenge 4}: technical capability to organize and orchestrate different AI agents to accomplish a single task.
\end{itemize}

\begin{figure}[t]
    \centering
    \includegraphics[width=0.85\textwidth]{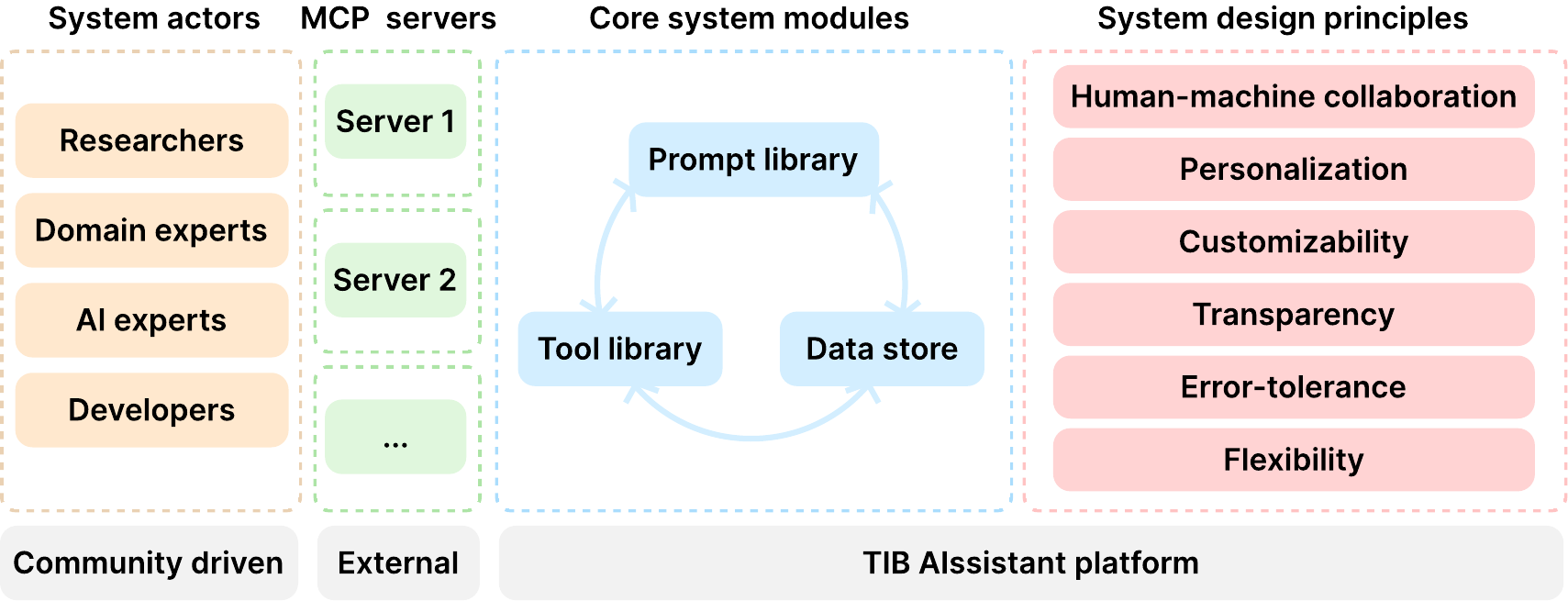}
    \caption{Proposed framework of AI-assisted research, highlighting system actors, external MCP (Model-Context Protocol) servers consisting of collections of tools, core system modules, and the system's design principles.}
    \label{fig:framework}
\end{figure}

In this work, we present our vision for an AI-supported, domain-agnostic research platform, named TIB AIssistant. \autoref{fig:framework} depicts the conceptual framework of our approach. The platform serves as a central repository for scholarly AI agents and their corresponding prompts. Additionally, the platform provides the scholarly tools necessary to accomplish research tasks. To serve researchers across various domains, we argue that the flexibility of the system is crucial for accommodating the diverse requirements and use cases arising from diverse research work. \autoref{fig:life-cycle} illustrates an example research life cycle within the TIB AIssistant. The user begins with using the TIB AIssistant to generate ideas, proceeds through the different phases of the life cycle (supported by various agents), and utilizes external tools as needed. The data store stores results from different agents and makes them available to other agents. 

\section{Vision of the TIB AIssistant}
Our vision for the TIB AIssistant is to empower researchers through an AI-supported, human-centered, and domain-agnostic platform that redefines the conduct of scholarly research. 
We envision a collaborative research environment where humans and machines co-create knowledge. 
Rather than aiming for full automation, the TIB AIssistant centers on \textit{human-machine collaboration}, enabling researchers to retain control, orchestrate processes, and critically evaluate AI-generated results throughout the research life cycle.

At the core of this vision is a \textit{flexible, modular, and transparent infrastructure} that facilitates AI integration without imposing rigid workflows. 
The TIB AIssistant is conceived as a lightweight yet powerful research hub where customizable AI agents, scholarly tools, and curated prompts work together seamlessly. 
Each element of the system — ranging from prompt libraries to external tool integrations — is designed to be interoperable, extensible, and openly accessible.

We aim to lower the barrier to AI adoption in academia by addressing the four key challenges faced by researchers:
i) understanding the scope of AI in domain-specific tasks,
ii) developing effective prompts and contextual inputs,
iii) integrating external scholarly tools into AI workflows,
iv) and coordinating diverse AI agents to perform complex research processes.

Inspired by principles from Integrated Development Environment (IDE) interfaces, the TIB AIssistant provides a research-friendly environment where users can initiate ideation and literature exploration, formulate research questions, iterate on methodologies, analyze and synthesize results, and ultimately author and refine scholarly publications. This vision is grounded in a set of foundational design principles: \textit{Personalization, Customizability, Trustworthiness and Transparency, Error-tolerance}, and \textit{Open science and community engagement}. Ultimately, the TIB AIssistant aspires to become a central AI hub for scholarly research, not just a tool, but an evolving community-driven platform that transforms how research is conceptualized, conducted, and communicated in the age of generative AI.

\section{Related Work}
The use of LLMs to support scholarly activities is a growing field. Existing approaches can be broadly categorized into single task assistance and those that offer a more integrated, multi-task framework.

\subsection{Single Task Assistants}
A significant body of work focuses on leveraging LLMs to streamline specific, often labor-intensive, components of the research process. For instance, the STORM approach \cite{shao2024assistingwritingwikipedialikearticles} provides a systematic approach to the research and pre-writing stages, which are the core of a literature review. 
Similarly, ResearchAgent~\cite{baek2025researchagentiterativeresearchidea} is an LLM-powered system focused on research idea generation by defining problems, proposing methods, and designing experiments. 
The system utilizes collaborative LLM-powered reviewing agents to refine these ideas iteratively based on feedback. Another specialized application of LLM agents is simulating complex scholarly interactions. The AgentReview framework~\cite{jin2024agentreviewexploringpeerreview}, for example, utilizes LLM-based agents to simulate the entire peer-review process, allowing for the study of its dynamics, including reviewer bias and the influence of author identity. Also, there are domain-specific tools such as \textit{Name2SMILES} (for converting molecule names to SMILES), \textit{ReactionPlanner} (for multi-step synthesis planning), \textit{PatentCheck} (for checking compound patent status), and \textit{SafetySummary} (for retrieving safety information) from the ChemCrow platform~\cite{branAugmentingLargeLanguage2024}, for research-related tasks in chemistry. 

\subsection{Multi-Task Assistants}
Moving beyond single-task applications, a growing number of initiatives aim to create more comprehensive, multi-faceted research assistants. These systems often integrate several capabilities to support researchers throughout their workflow. Paper Copilot~\cite{lin2024papercopilotselfevolvingefficient}, for example, functions as a personalized academic assistant that maintains a real-time updated database of research papers. It can derive a user's research profile, analyze the latest trending topics, and provide advisory services, thereby combining multiple support functions into a single system. Furthermore, there is The AI Scientist~\cite{lu2024aiscientistfullyautomated}, a framework designed for fully automated, open-ended scientific discovery. This system represents a significant step towards end-to-end automation by autonomously performing a sequence of research tasks. Starting with ideation based on existing literature, followed by experimentation, and finally, the paper authoring. This results in a complete manuscript in LaTeX, which LLM agents also review. 

While these approaches demonstrate the potential of full automation, their rigid, pipeline-driven nature highlights several challenges that our vision for the TIB AIssistant aims to address. The emphasis on a fully autonomous process limits the role of the researcher, contrasting with our core principle of human-machine collaboration, where the human expert orchestrates and validates each step. The fixed workflow within the AI assistant also fall short of our goal for a customizable and flexible platform, where users can modify prompts, select different LLMs, and integrate their tools across various disciplines. 

\begin{figure}[t]
    \centering
    \includegraphics[width=0.9\textwidth]{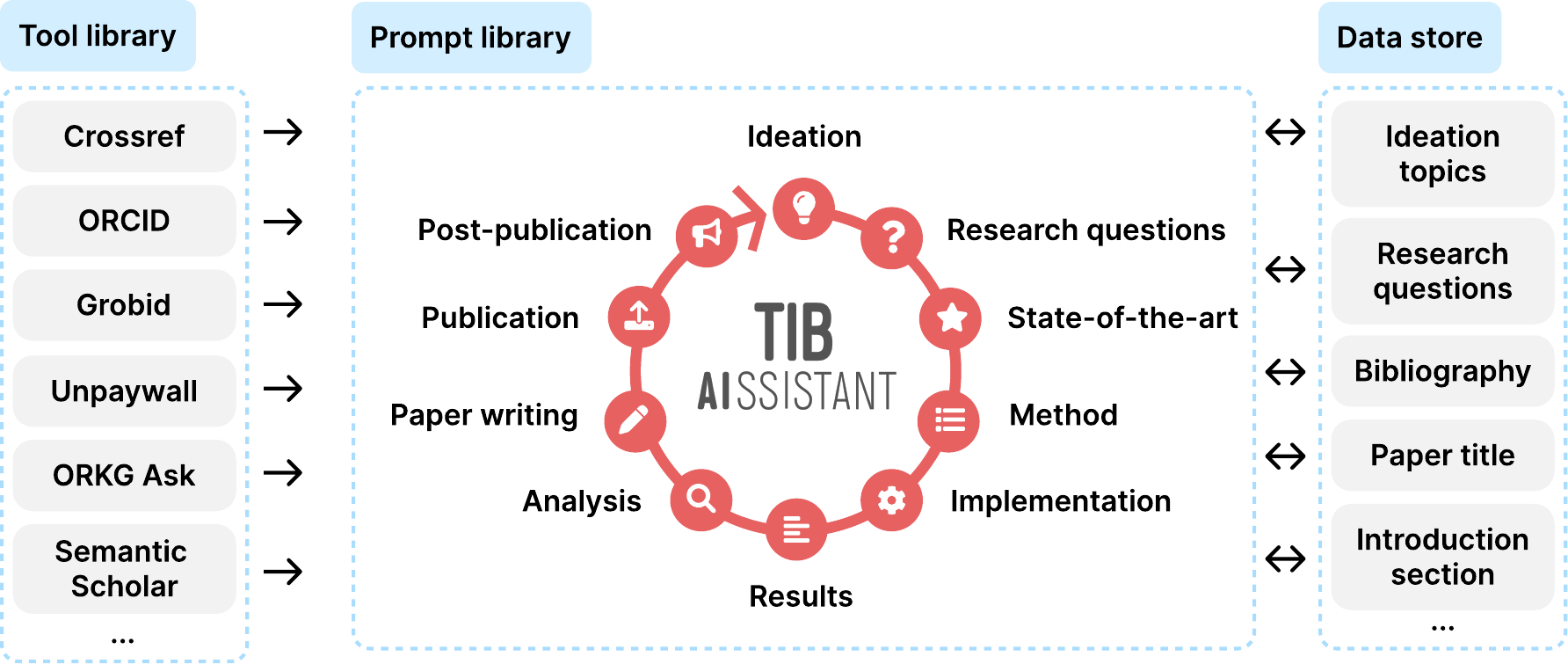}
    \caption{Example research life cycle starting from ideation until post-publication. The listed items provide concrete examples of the three core system modules, as shown in \autoref{fig:framework}. External tools are listed, along with items stored within the data store.}
    \label{fig:life-cycle}
\end{figure}

\section{Framework for AI-Assisted Research}
We now describe the main components that form the foundational framework of our envisioned approach toward AI-supported research. The framework's concepts are discussed next and summarized in~\autoref{fig:framework}.
The platform provides an integrated environment for researchers, much like an IDE for software developers. It consists of a Graphical User Interface (GUI) allowing users to interact with various agents. We consider this platform as a lightweight wrapper that integrates the different components listed below. If existing approaches or tools are available, the platform should implement these services instead of attempting to replicate their functionalities.

\subsection{Core System Modules}

\paragraph{Prompt Library}
The Prompt Library is a collection of system prompts tailored toward specific tasks of the research life cycle. A list of prompts minimizes the need for researchers to create their own prompts, often relying on time-consuming trial-and-error. The prompt library enables researchers to share their approaches with others easily. In addition to the prompt, metadata must be assigned to the prompt to indicate what task is addressed (e.g., research question formulation, finding related literature, etc.). Multiple variants of prompts can exist for the same task, thus providing alternatives in case a prompt does not produce the expected result. Finally, users should be able to see and modify prompts when using them within the platform. 

\paragraph{Tool Library}
The Tool Library integrates external services into the platform. This makes it possible to connect the platform to external tools, for example, to fetch additional publication data from Crossref and ORCID, or to fetch related work via Semantic Scholar~\cite{kinney2025semanticscholaropendata}, ORKG~\cite{Auer.2025}, or ORKG Ask~\cite{oelen2024orkg}. Tools are called automatically, where the LLM decides, based on the description of the tool and the user's input, whether a tool should be called or not. To ensure tools can be added dynamically, a Model Context Protocol (MCP)~\cite{modelcontextprotocolIntroductionModel} can be used. MCP provides a standardized approach to provide external access to LLMs. In this case, we are specifically interested in enabling calling tools. This enables users to integrate existing scholarly MCP servers, allowing them to access a range of scholarly tools quickly and easily. For developers, it is possible to set up an MCP server to make their tools available to the platform. To facilitate external tool integration via MCP servers, we plan to develop an MCP GUI tool, enabling the easy setup of an MCP server for tools that utilize REST endpoints.

\paragraph{Data Store}
The ability of different agents to communicate with each other can be accomplished via a centralized data store. Compared to keeping all generated content in the context of the LLM, this approach has several benefits: the constraint of context window size is less problematic since the data is stored in a separate store, and the agents are self-contained. They can be used in isolation, making it easier to reason about what is happening within the agent. The data store can be a relational database, storing specific data under a predefined key (e.g., research questions, bibliography, etc.). When necessary, the database can be accessed, and the respective content is added to the context. A more advanced approach could provide the LLM with a tool that allows it to access the database, enabling it to read from and write to the database automatically. 

\subsection{System Design Principles}

\paragraph{Human-Machine Collaboration}
In the spectrum between full automation of research and humans doing most of the required tasks, a hybrid approach where humans and machines collaborate offers the best of both worlds. In such a hybrid approach, the human researcher primarily orchestrates, directs, and reviews AI-supported processes. In this model, researchers have control at all times and review and evaluate the output created by the AI at each step. This means that user interfaces must be designed to go beyond the conventional conversational prompt-response style, allowing users to modify intermediary results and decide when to proceed to the next step. However, due to the conversational setup, processes such as iterative refinement enable users to refine AI-generated responses through a human-in-the-loop approach further. We argue that this type of control is essential for an AI-supported research assistant to be both useful and adopted by researchers. Therefore, the interface should not aim to provide an automated research pipeline, but instead offer a highly customizable environment that researchers can use to integrate AI support into their workflows. 

\paragraph{Personalization}
A sophisticated memory system is crucial for enhancing the platform's effectiveness through deep personalization. This feature enables dynamic context engineering~\cite{mei2025surveycontextengineeringlarge}, a process where only the most relevant information for a given task is selectively retrieved or fetched and then provided to the LLMs, optimizing both relevance and computational efficiency. Beyond concrete tasks, the memory system constructs a user profile that learns about domain expertise, stylistic preferences, formatting conventions, and preferred terminology over time. Moreover, the memory component maintains a comprehensive record of the user's research history. This historical knowledge would empower the assistant to guide and coordinate various (sub-)agents, ensuring their outputs are consistent and aligned with the overarching user preferences.

\paragraph{Customizability}
The platform should be domain-agnostic and sufficiently flexible to support different workflows. Users should be able to customize the platform to support their use cases. This begins with the prompts, where users can try out different variants and edit them as needed. Additionally, the list of tools that the LLM can execute during a chat session should be modifiable to limit the scope of available tools and better direct the LLM in selecting the appropriate tool. As previously mentioned, the platform should serve as a lightweight tool connecting different services. The ability to customize the interface is therefore crucial to support a variety of use cases. 

\paragraph{Transparency and Trustworthiness}
For transparency and reproducibility reasons, for all generated artifacts, provenance data has to be recorded, capturing the creators, model name, and model version, system and user prompts, invoked tools, etc. To provide complete transparency, this provenance data should be published alongside the research paper as research data. We envision this data to be published in a machine-readable format, for example, via RO-Crates~\cite{dreyes2022packaging}, to facilitate machine actionability. These aspects will also help users verify the accuracy and originality of the generated content.

\paragraph{Error-Tolerance}
With an error-tolerant interface, we ensure that users remain in control and can modify any data generated by the AI. This means that all messages, including system and user messages, as well as the generated data, should be modifiable by the user. Additionally, the user should be able to see which data is provided as input when the LLM calls a tool. This helps determine whether the tool was called as expected. Based on our previously mentioned aspect of Human-Machine Collaboration, the user can decide which data to use and which to discard. Regarding external tool callings, since we do not control these services ourselves, we should assume that these services may respond differently than expected (e.g., because of a temporary issue, rate limiting, or changes in API specifications). In such cases, the AIssistant should be error tolerant by gracefully handling such errors.

\paragraph{Flexibility}
We envision a centrally hosted platform. Users of the platform should have the flexibility to choose the models and LLM providers they prefer. This also serves the purpose of selecting providers based on geographical locations, legal requirements, and privacy concerns, among other factors. 
Smaller models can handle simple tasks, while larger models can execute more complex tasks. A limited number of free tokens can be provided to each user per day, which can be managed through an authentication system. A Bring Your Own Key (BYOK) approach can be used to allow users to use as many tokens as necessary. As an alternative, users should also be able to run the platform locally on their computers, as the source code is publicly available. 

\paragraph{Community Driven}
We aim to develop the platform in collaboration with academics from various domains. As there are virtually unlimited use cases for AI-supported research, the platform relies on community contributions to create prompts and add tools. An additional way to contribute is by adding functionalities to the platform itself. 

\section{Conclusion and Outlook}

In this work, we laid the foundation for an AI-supported research platform. 
To address challenge 1, we propose a library of prompts that showcases to researchers the types of tasks that can be performed by LLMs and the tools required for these tasks. Challenge 2 is addressed by reducing the need to engineer prompts and by automatically providing the required context for a prompt through the data store. Challenge 3 is tackled by integrating external scholarly tools via tool calls and external MCP servers. Finally, challenge 4 is addressed by providing predefined research life cycle workflows, where different agents can interact with each other by means of sharing data via the data store. 

A key aspect of our approach is the community effort to create and curate prompts and tools in such a manner that they are helpful for other users in the community. We aim to provide the technical means to make this possible. This includes the addition of community features, such as voting for helpful prompts and sharing custom workflows with other users. In the end, we envision that the prompt library will contain different versions of prompts aiming to accomplish the same research task. This follows the assumption that there is no one-size-fits-all approach, but that different prompt variants are helpful for different use cases. 

An initial prototype is implemented, integrating the proposed framework concepts into a workable research life cycle. A demonstration of this approach is published~\cite{oelen2025aissistantdemo}. \autoref{fig:life-cycle} depicts parts of the prototype implementation. Only domain-agnostic assistants are implemented in the prototype (i.e., ideation, research questions, state-of-the-art, paper writing). Outputs from, for example, ideation, are utilized by other assistants, such as when formulating research questions and during paper writing. Furthermore, the tools and data store items listed in the figure are also implemented. The prototype implementation shows the feasibility of our approach and demonstrates how the core system modules are integrated. The source code is available online.\footnote{\url{https://gitlab.com/TIBHannover/orkg/tib-aissistant/web-app}}. We plan to further develop the prototype into a publicly available online service, where researchers can get started with AI-assisted research. 

\begin{acknowledgments}
We thank our colleague Markus Stocker for his valuable comments in reviewing this paper. This work was co-funded by NFDI4DataScience (ID: 460234259) and by the TIB Leibniz Information Centre for Science and Technology.
\end{acknowledgments}


\section*{Declaration on Generative AI}
During the preparation of this work, the authors utilized ChatGPT, Gemini, and Grammarly to draft content, enhance content, paraphrase and reword, improve writing style, and Perform Grammar and spelling checks. After using this service, the authors reviewed and edited the content as needed and take full responsibility for the publication's content. 

\bibliography{refs}

\appendix

\end{document}